\documentclass[lettersize,journal]{IEEEtran}
\usepackage{amsmath,amsfonts}
\usepackage{algorithmic}
\usepackage{algorithm}
\pdfoutput=1
\usepackage{array}
\usepackage[caption=false,font=normalsize,labelfont=sf,textfont=sf]{subfig}
\usepackage{textcomp}
\usepackage{stfloats}
\usepackage{url}
\usepackage{verbatim}
\usepackage{graphicx}
\usepackage{cite}
\usepackage{booktabs}
\usepackage{multirow}
\usepackage{balance}
\usepackage{color}
\begin{document}

\title{Scale Guided Hypernetwork for Blind Super-Resolution Image Quality Assessment}

\author{Jun Fu}

\markboth{Journal of \LaTeX\ Class Files,~Vol.~14, No.~8, August~2021}%
{Shell \MakeLowercase{\textit{et al.}}: A Sample Article Using IEEEtran.cls for IEEE Journals}


\maketitle
 
\begin{abstract}
With the emergence of image super-resolution (SR) algorithm, how to blindly evaluate the quality of super-resolution images has become an urgent task. However, existing blind SR image quality assessment (IQA) metrics merely focus on visual characteristics of super-resolution images, ignoring the available scale information. In this paper, we reveal that the scale factor has a statistically significant impact on subjective quality scores of SR images, indicating that the scale information can be used to guide the task of blind SR IQA. Motivated by this, we propose a scale guided hypernetwork framework that evaluates SR image quality in a scale-adaptive manner. Specifically, the blind SR IQA procedure is divided into three stages, i.e., content perception, evaluation rule generation, and quality prediction. After content perception,  a hypernetwork generates the evaluation rule used in quality prediction based on the scale factor of the SR image. We apply the proposed scale guided hypernetwork framework to existing representative blind IQA metrics, and experimental results show that the proposed framework not only boosts the performance of these IQA metrics but also enhances their generalization abilities. Source code will be available at \textcolor{red}{https://github.com/JunFu1995/SGH}.
\end{abstract}


\section{Introduction}
\IEEEPARstart{I}{mage} super-resolution (SR) is an effective technique to construct a high-definition (HD) image with finer details from one or multiple low-resolution images. The past two decades have witnessed the bloom of image SR algorithms with a broad of applications including video surveillance~\cite{pais2014super}, video streaming~\cite{dasari2020streaming,deepstream}, free-viewpoint television~\cite{garcia2012super,jin2015virtual}, etc. However, how to reasonably evaluate the quality of SR images to compare and optimize SR methods remains challenging. 

The most credible approach for SR image quality assessment (IQA) is the subjective test, where the mean of scores rated by subjects is regarded as the visual quality of SR images. However, the subjective test is time-consuming, expensive, and cannot be applied to real-world SR optimization systems since it requires extensive human interaction. Consequently, objectively evaluating the visual quality of SR images emerges as time requires, which is consistent with subjective evaluation but not human-in-loop. 

\begin{figure}[htbp]
	\centering
	\begin{tabular}{cc}
		\includegraphics[width=\linewidth]{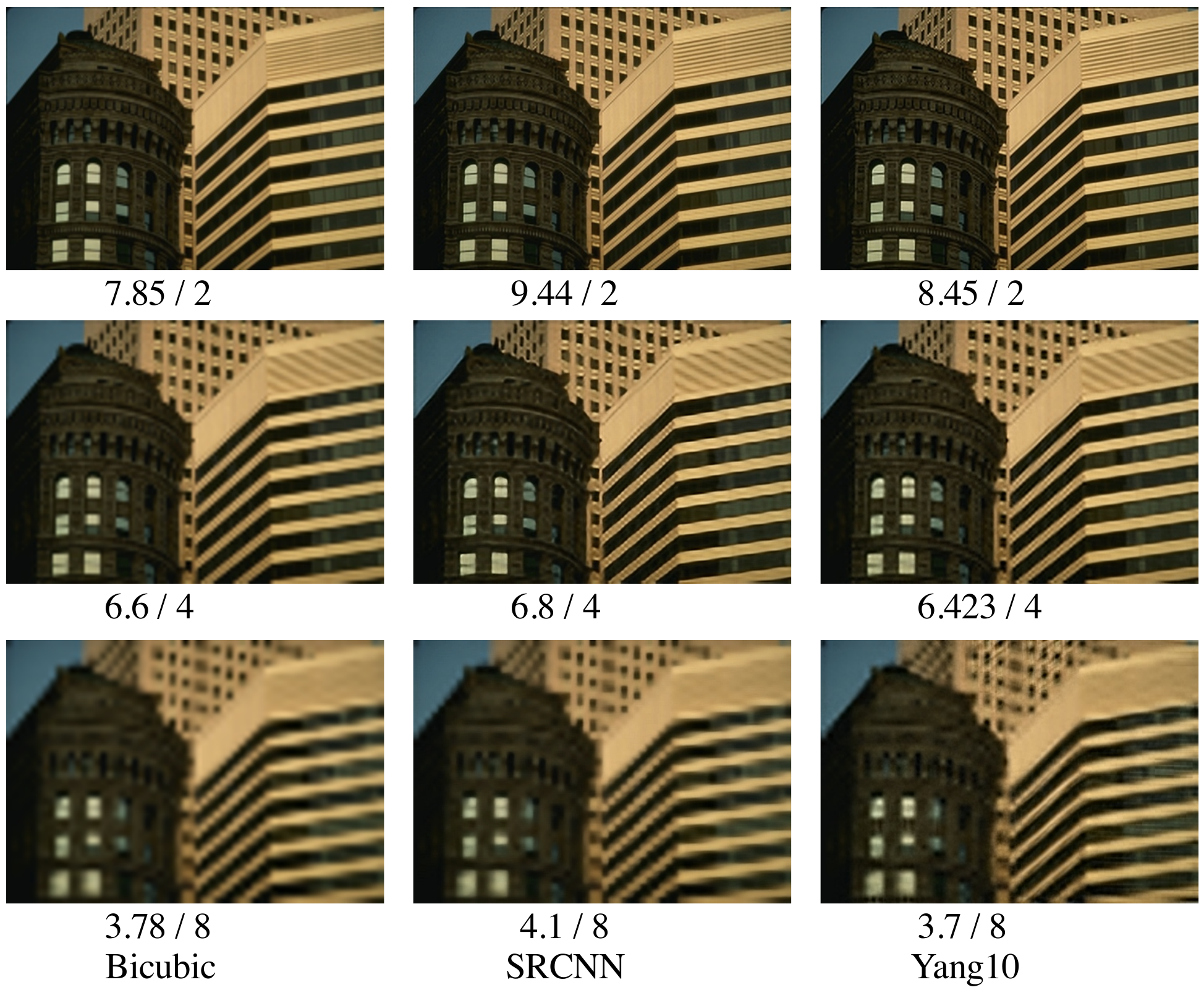} &
	\end{tabular}
	\caption{Examples of SR images generated by Bicubic, SRCNN~\cite{dong2014learning}, and Yang10~\cite{yang2010image}. The text under each SR image contains the subjective quality score and the scale factor of the SR image.}
	\label{fig:scaleDiff}
\end{figure}

Existing objective IQA metrics can be classified into three categories: full-reference (FR) metrics which require the original uncorrupted image as a reference, reduced-reference (RR) metrics which require partial information about the reference image, and no-reference (NR) metrics (also known as blind metrics) which do not need any reference information. In practical SR systems, the original HD image is generally unavailable. As a result, most FR  and RR metrics might not be used for SR IQA. Thus, it is highly desired to develop effective blind IQA metrics to evaluate the visual quality of SR images. 

In general, there are various spatial artifacts (such as blurring and blocking effects) in the generated SR images compared to the original HD images, of which the loss of high-frequency information is the main cause. As a result, earlier blind SR IQA approaches exploit statistical features in spatial and frequency domains to quantify artifacts of SR images. However, these methods~\cite{yeganeh2015objective, ma2017learning, deepSRQ, SRQC} typically involve manual feature engineering, which is a laborious and error-prone process. To alleviate this problem, Bare et al.~\cite{bare2018deep} propose the first convolutional neural network-based blind SR IQA metric, which automatically learns the mapping from image to quality. With the success of the first deep learning-based blind SR IQA measure, researchers shift their focus from feature engineering to neural architecture engineering. In the past few years, various neural architecture strategies including deeper network architecture~\cite{lu2022deep}, dual-branch framework~\cite{zhao2021learning,liu2022textural}, and visual attention mechanisms~\cite{JCSAN}, have been proposed to further improve the performance. 

\begin{figure*}[htbp]
	\centering
	\begin{tabular}{ccc}
		\includegraphics[width=0.31\linewidth]{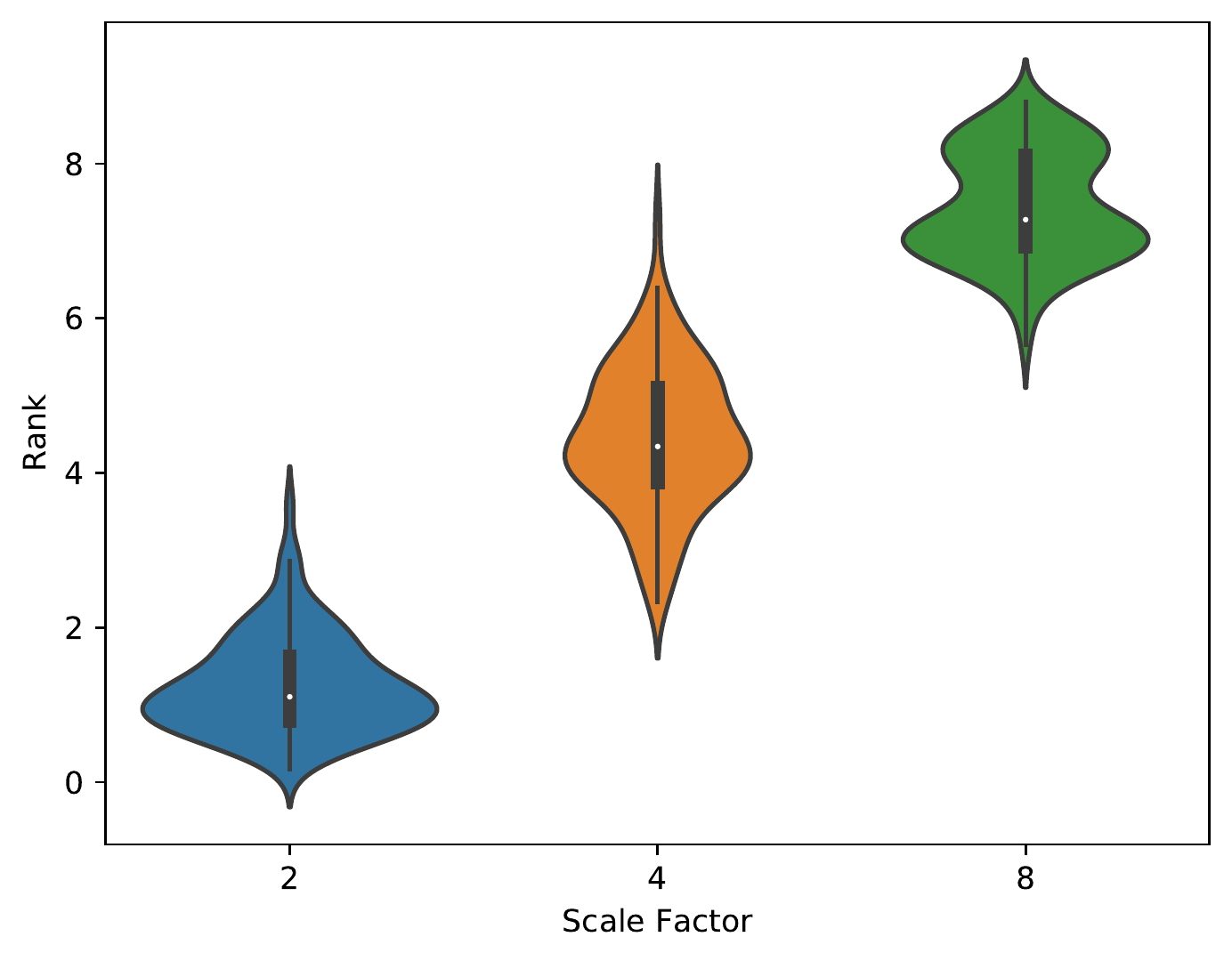} & \includegraphics[width=0.31\linewidth]{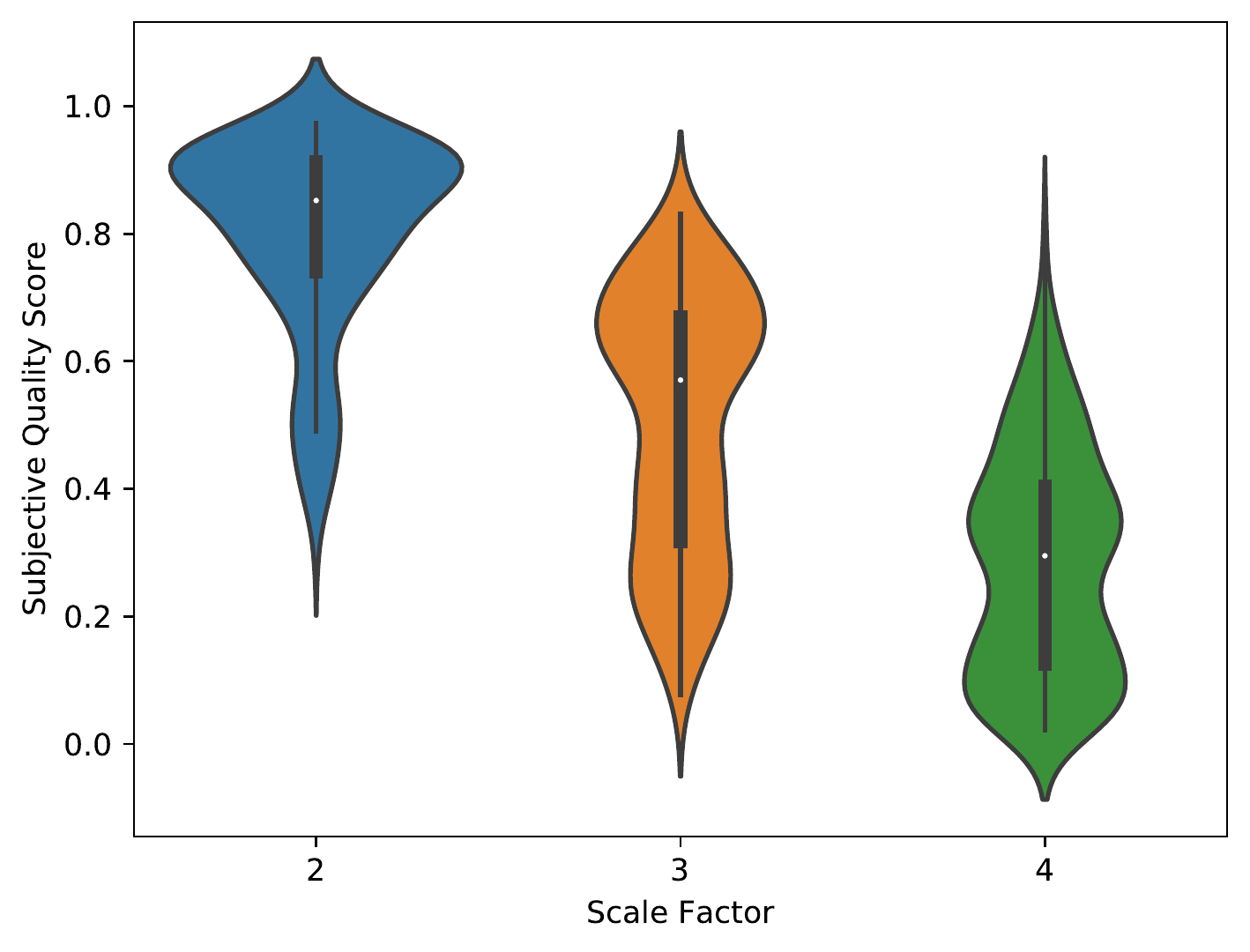} &
		\includegraphics[width=0.31\linewidth]{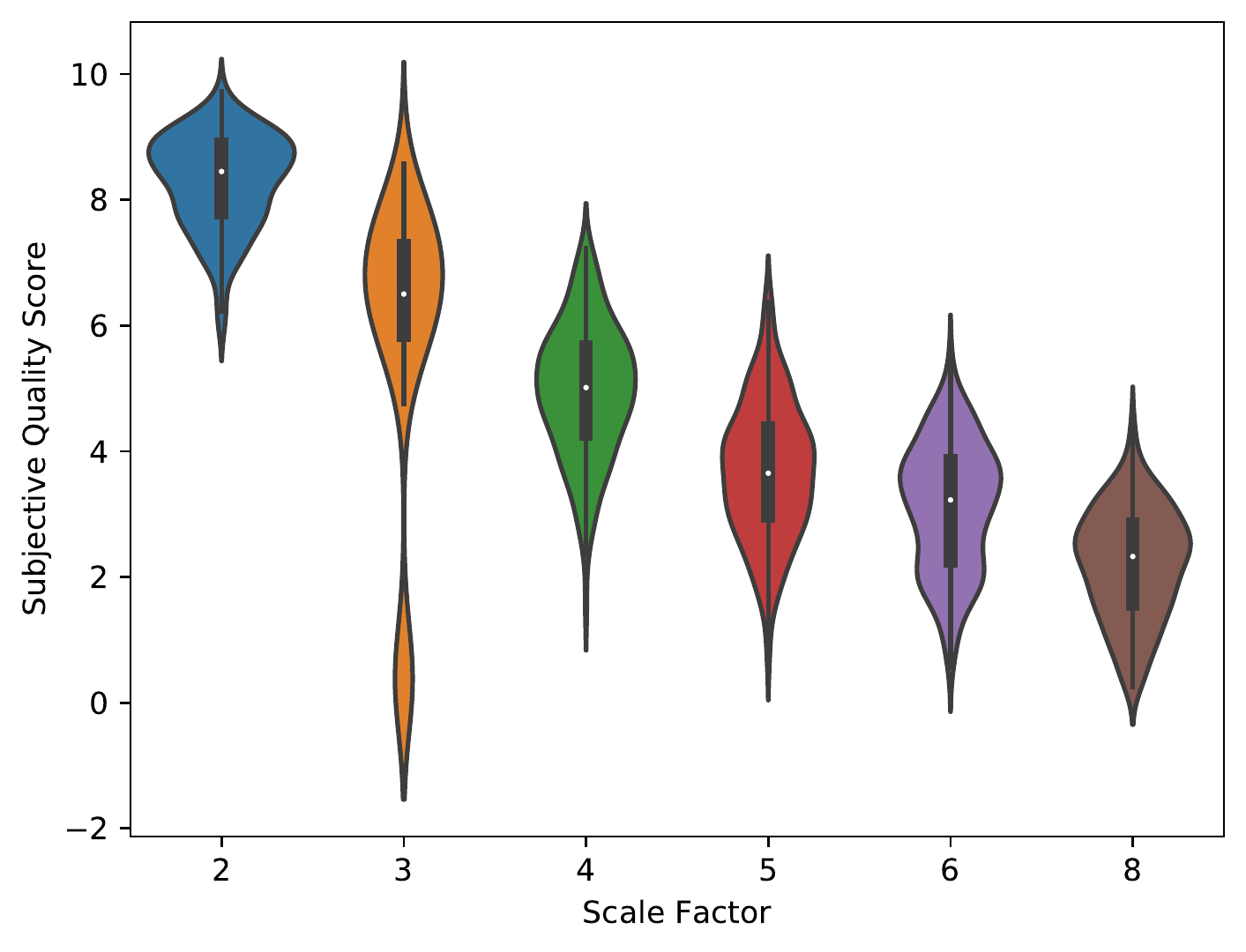} \\ 
		 (a) & (b) & (c)\\ 
	\end{tabular}
	\caption{Distribution of subjective quality score under different scale factors. (a) On the Waterloo dataset; (b) On the QADS dataset; (c) On the CVIU dataset. }
	\label{fig:scale}
\end{figure*}

The aforementioned blind SR IQA methods focus on learning the mapping from image to quality, but pay no attention to available quality-aware priori, i.e., scale factors of SR images. As shown in Fig.~\ref{fig:scaleDiff}, the visual quality of the SR images generated by the three SR algorithms and their subjective quality scores decrease significantly with the increase of the scale factor. This indicates that the scale information of the SR image is highly correlated with its subjective quality score. In other words, the scale information is useful for the task of SR IQA. However, how to effectively use scale information to guide the quality evaluation of SR images remains open. 

One vanilla method is to concatenate the scale information and the image in the feature space and map the fused features to the quality score. Unfortunately, such an approach has limited performance, which is proven in the section of the experiment. To address this issue, we propose a scale guided hypernetwork framework, which adopts distinct evaluation rules for SR images with different scale factors. Specifically, the proposed framework divides the blind SR IQA procedure into three stages, i.e., content perception, evaluation rule generation, and quality prediction. After content perception, the scale-adaptive evaluation rule is generated by a hypernetwork, and then used for quality prediction.

Our major contributions are as follows:
\begin{itemize}
	\item Up to our best knowledge, this paper is the first to reveal that the available scale information can be used to guide the task of blind SR IQA. 
	\item We present a scale guided hypernetwork framework, which fills the blank in scale-guided blind SR IQA.
	\item We conduct sufficient experiments to verify the effectiveness and superiority of the proposed method. 
\end{itemize}

The remainder of this paper is organized as follows. Section II introduces the related work. Section III explains the motivation of this paper. Section IV elaborates the proposed scale guided hypernetwork. Section V provides the experimental results. Finally, the paper is concluded in Section VI.

\section{Related Work}
\subsection{Super-Resolution Image Quality Assessment}
Since we design a blind IQA metric for image super-resolution, we only review existing blind SR IQA methods. Generally, existing blind super-resolution image quality assessment (SR IQA) metrics could be classified into two main categories: handcrafted feature-based methods~\cite{yeganeh2015objective, ma2017learning, deepSRQ, SRQC} and deep learning-based ones~\cite{bare2018deep,fang2018blind,lu2022deep,zhao2021learning,liu2022textural,JCSAN}. 

In the first category, researchers are devoted to regressing the quality score of the SR image based on handcrafted features. Yeganeh et al.~\cite{yeganeh2015objective} manually design three statistical characteristics for blind SR IQA: frequency energy falloff statistics, dominant orientation statistics, and spatial continuity statistics. Ma et al.~\cite{ma2017learning} blindly evaluate SR images based on their local frequency, global frequency, and spatial discontinuity. Zhou et al.~\cite{deepSRQ} map the texture and structure of the SR image to the quality score. Greeshma et al.~\cite{SRQC} build a super-resolution quality criterion based on the estimation of fuzzy gradient profiles and multi-scale energy bands. Although these handcrafted features show promising results, their generalization abilities are typically limited. 

In the second category, researchers aim to automatically learn the mapping from image to quality. Bare et al.~\cite{bare2018deep} and Fang et al.~\cite{fang2018blind}  model the mapping function through shallow neural networks, which only consist of several convolutional layers. In order to achieve better modeling ability, various neural architecture design strategies are proposed. 
Lu et al.~\cite{lu2022deep} deepen the depth of IQA models, while Zhao et al.~\cite{zhao2021learning} and Liu et al.~\cite{liu2022textural} increase the width of IQA models, i.e., adopting dual-branch architectures. Besides, Zhang et al.~\cite{JCSAN} introduce channel-spatial attention mechanisms into IQA models. However, these methods are still performance limited since they only consider the image features while ignoring the available quality-aware priori, i.e., the scale information. In this paper, we first verify that the scale information is highly correlated with subjective scores of SR images through statistical analysis, and then present the first scale guided hypernetwork framework to effectively exploit the scale information for blind SR IQA.

\subsection{Hypernetwork}
The hypernetwork~\cite{ha2016hypernetworks} is a neural network that generates the parameters of another network, which has been widely used in neural architecture search~\cite{zhang2018graph}, light field compression~\cite{zhu2022minl}, and general image quality assessment~\cite{su2020blindly}. Here, we only discuss the related work ~\cite{su2020blindly} in detail. In the hypernetwork-based general IQA method~\cite{su2020blindly}, the evaluation rule, i.e., the weights and biases of fully-connected layers used in the stage of quality prediction, is adaptively generated according to image semantics. While this method shows impressive generalization capabilities for general IQA tasks, it is not optimal to apply it directly to SR IQA tasks. In this paper, we design a scale guided hypernetwork framework specifically for SR IQA tasks.

\section{Motivation}
In this section, we qualitatively and quantitatively verify the significant impact of scale factors on subjective quality scores of SR images, which motivates us to develop a scale guided hypernetwork framework for blind SR IQA.

\subsection{Qualitative Analysis}
We divide SR images into groups based on their scale factors and then visualize the distribution of subjective quality scores in each group through violin plots. The results are presented in Fig.~\ref{fig:scale}. As shown, each group has a different distribution of subjective quality scores. In addition, the average MOS (Rank) of each group decreases (increases) with the increase of the scale factor. Therefore, we conclude that the scale factor could provide us with some clues about the visual quality of SR images.

\subsection{Quantitative Analysis}
\begin{table}[htbp]
	\centering
	\renewcommand\arraystretch{1.2}
	\caption{Alexander-Govern approximation tests.}
	\begin{tabular}{|c|c|c|}
		\hline
		Datasets & Statistic value & $p$ value \\
		\hline 
		CVIU~\cite{ma2017learning} & 84.67 & 3.23e-14 \\
		\hline 
		Waterloo~\cite{yeganeh2015objective} & 49.48 & 7.09e-11\\
		\hline 
		QADS~\cite{zhou2019visual} & 89.83 & 8.67e-11 \\
		\hline
	\end{tabular}
	\label{tbl:A-test}
\end{table}
Based on the results of the qualitative analysis, it is natural to ask whether the differences among all groups are statistically significant. Therefore, we conduct a one-way Analysis of Variance (ANOVA). Specifically, ANOVA quantifies the differences in means between groups:
\begin{equation}
g, p = G(m_0, ...,m_i,..., m_n),
\end{equation}
where $m_i$ represents the average MOS (Rank) of all SR images with the $i$-th scale factor, and $G$ produces the statistic value $g$ and the probability $p$ according to $\{m_i\}^{n}_{i=0}$. When the probability $p$ is less than or equal to the significance level $\alpha$, we can conclude that all groups are statistically significantly different. The significance level $\alpha$ is commonly set to 0.05, and $G$ employs the Alexander-Govern approximation test~\cite{alexander1994new} since it does not assume homoscedasticity. In this paper, we do not consider differences introduced by SR methods. Therefore, we calculate the statistic value $g$ and the probability $p$ for each SR method and report the average results in Table~\ref{tbl:A-test}. As shown, the probability $p$ is far smaller than the significance level with a value of 0.05 across three datasets:
\begin{equation}
p << \alpha.
\end{equation}
Therefore, we conclude that the scale factor has a statistically significant influence on the visual quality of the SR image. In other words, the scale factor is useful for the task of SR IQA.

\begin{figure*}[htbp]
	\centering
	\includegraphics[width=\linewidth]{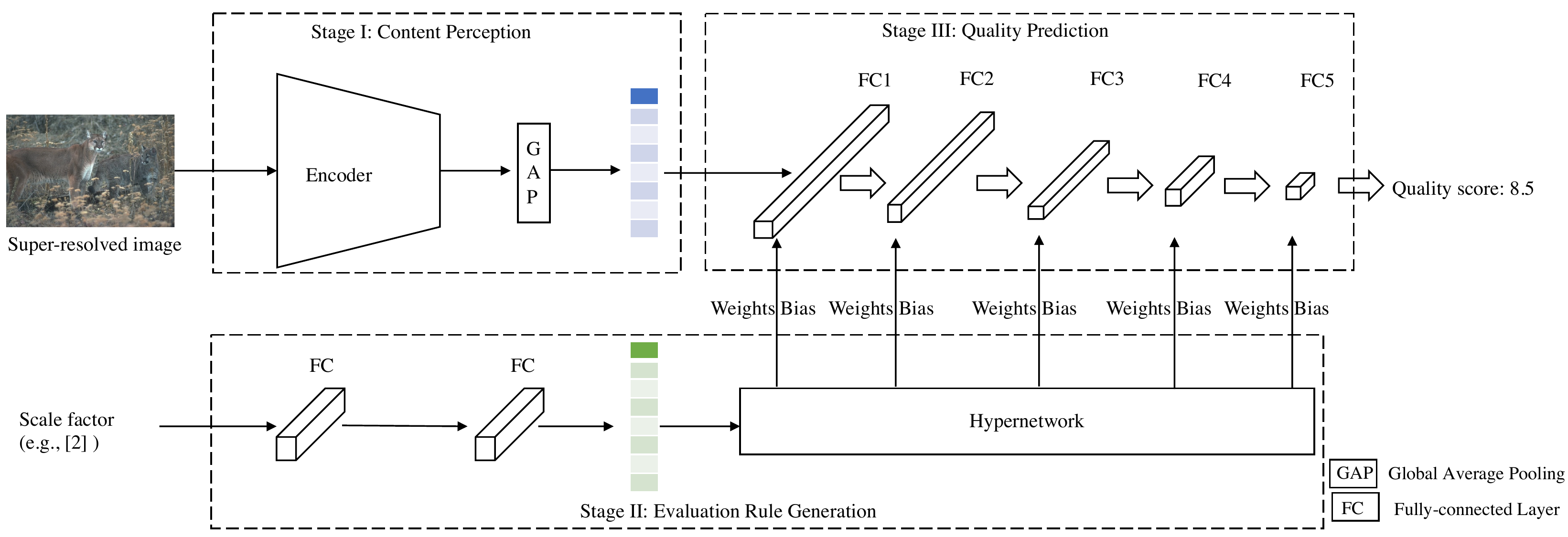} 
	\caption{The diagram of scale guided hypernetwork framework.}
	\label{fig:framework}
\end{figure*}

\section{Method}
\subsection{Problem Formulation}
Given a low-resolution image $I_{L}$ and its super-resolution version $I_{H}$, we aim to predict the visual quality of the high-definition image $I_{H}$. Generally, this task is formulated as a regression problem:
\begin{equation}
q = f(I_{H};\theta),
\label{Eq:goal}
\end{equation}
where the function $f$ with parameters $\theta$ maps the super-resolution image $I_{H}$ into the quality score $q$, and parameters $\theta$ are shared among all super-resolution images. 

According to the aforementioned analysis, we know that it is necessary to take the scale information into account in the task of blind SR IQA. However, how to effectively use scale information to guide the quality evaluation of SR images remains unexplored. One vanilla method is to concatenate the scale information and the high-definition image in the feature space and predict the quality score based on the fused features. Unfortunately, such an approach has limited performance, which will be disscussed in detail in the section of the experiment. To address this issue, we proposed a scale guided hypernetwork framework, which uses the scale information to generate scale-aware quality predictor. Therefore, we reformulate Eq.~\ref{Eq:goal} as follows: 
\begin{equation}
\begin{split}
q &= f(I_{H};\theta_{s_{I_{H}}}), \\
\theta_{s_{I_{H}}} &=  h(s_{I_{H}};\beta),
\label{Eq:Newgoal}
\end{split}
\end{equation}
where the function $f$ predict the quality score $q$ with image-aware parameters $\theta_{I_{H}}$, and the function $h$ with parameters $\beta$ generates $\theta_{s_{I_{H}}}$ based on the scale factor $s_{I_{H}}$. It is worth noting that the scale factor $s_{I_{H}}$ is not explicitly provided, but can be implicitly derived as follows: 
\begin{equation}
s_{I_{H}} = \frac{W_{I_{H}}} {W_{I_{L}}},
\end{equation}
where $W_{I_{H}}$ and $W_{I_{L}}$ are the width of $I_{H}$ and $I_{L}$, respectively.

\subsection{Framework Overview}
As shown in Fig.~\ref{fig:framework}, the proposed scale guided hypernetwork consists of three stages, i.e., content perception, evaluation rule generation, and quality prediction. In the first stage, we extract distinguishable features from the input super-resolution image. Then, we feed the scale factor into the hypernetwork for generating the evaluation rule, i.e., weights and biases of fully-connected layers. Finally, we map extracted distinguishable features into the quality score through the generated evaluation rule. The details of these three stages are introduced in sequence.  

\subsection{Content Perception} 
Generally, SR images suffer from severe high-frequency information loss, which leads to hierarchical distortions such as low-level texture damage and high-level structure degradation. Therefore, in the task of SR IQA, it is necessary to discover representative features that reflect these visual distortions. Considering that manually designing features is time-consuming, we resort to automatic feature engineering. Specifically, we use a learning-based encoder to capture the distortion-aware information:
\begin{equation}
F = f_e(I_{H}; \theta_e),
\end{equation}
where the function $f_e$ typically represents classification model with the pretrained parameters $\theta_e$, and $F$ is the resulted 3D feature maps. Since the stage of quality prediction takes the 1D vector as input, we apply a globally average pooling and a flattened operation over $F$: 
\begin{equation}
V = flatten(GAP(F)).
\end{equation}
It is worth noting that the improved strategies of feature extraction such as multi-scale representations could be easily incorporated into our framework.

\subsection{Evaluate Rule Generation} 
In the framework of hypernetwork-based IQA, the evaluation rule, i.e., the weights and biases of fully-connected layers used in the stage of quality prediction, is adaptively generated according to image semantics. While this framework exhibits impressive generalization capabilities in the general IQA tasks, directly applying it into SR IQA tasks is not optimal. Considering that the scale factor has a statistically significant impact on the visual quality of SR images, we generate the evaluation rule according to the scale factor of SR images.

Assume $N_q$ fully-connected layers are used in the stage of quality prediction, and the parameters of the $j$-th layer are the weights $w_j$ and the biases $b_j$, where $j\in\{0,1,...,N_{q-1}\}$. Then, the parameters of each layer are independently generated based on the scale representation $S$: 
\begin{equation}
w_j = FC_{j,1}(S), \quad b_j = FC_{j,2}(S),
\end{equation}
where $FC_{j,1}$ and $FC_{j,2}$ are fully-connected layers. Here, the scale representation $S$ is extracted as follow:
\begin{equation}
S = f_s(S_{I_H}), 
\end{equation} where $f_s$ consists of two fully-connected layers. 

\subsection{Quality Prediction} 
After the stage of evaluation rule generation, we use the generated parameters to map the extracted features into the quality score:
\begin{equation}
q = f_t(V; \{w_j, b_j\}^{N_q-1}_{j=0}),
\end{equation}
where $q$ is the predicted score. The  mean absolute error is employed as the loss function:
\begin{equation}
Loss = \frac{1}{M} \sum_{m=1}^{M} \lvert q_m - Q_m\rvert,
\end{equation}
where $Q_m$ represents the ground-truth quality score of the $m$-th SR image, and $M$ indicates the number of SR images in a mini-batch.

\begin{table*}[htbp]
	\centering
	\renewcommand\arraystretch{1.1}
	\caption{Details of three popular SR IQA datasets.}
	\begin{tabular}{ccccccc}
		\hline
		datasets & LR Images&  SR Images & Number of SR methods& Scale factors & Labels & Label Range  \\
		\hline 
		Waterloo & 13 & 312 & 8 & \{2,4,8\} & Rank & [1,5]\\ 		\hline 
		
		QADS & 60 & 980 & 21 & \{2,3,4\} & MOS & [0, 1] \\ 		\hline 
		
		CVIU & 30  & 1620 & 9 &  \{2,3,4,5,6,8\} & MOS & [0, 10]  \\

		\hline
	\end{tabular}
	\label{tbl:dataset details}
\end{table*}
\begin{table*}[!tbhp]\small
	\renewcommand\arraystretch{1.1}
	\centering
	\caption{Single-dataset performance results of Scale guided hypernetwork (SGH) framework.}
	\label{tab:basemodel}
	\begin{tabular}{l|c|ccc|ccc|ccc}
		\toprule
		\multirow{2}{*}{Base Model}                                                                  & \multirow{2}{*}{Framework} & \multicolumn{3}{c|}{Waterloo}    & \multicolumn{3}{c|}{QADS} & \multicolumn{3}{c}{CVIU} \\ \cline{3-11} 
		&      & SRCC   &  PLCC & KRCC    & SRCC        &  PLCC & KRCC          & SRCC       &  PLCC & KRCC     \\ \hline
		
		\multirow{2}{*}{CNNIQA~\cite{CNNIQA}}  & Original & 0.8717&0.8829&0.7318&0.8169&0.7756&0.6411&0.6789&0.6750&0.5154\\    

		& SGH & 0.9360&0.9731&0.8040&0.8906&0.8880&0.7300&0.9104&0.9281&0.7525\\ \hline     
		
		\multirow{2}{*}{JCSAN~\cite{JCSAN}}  & Original & 0.8839&0.8696&0.7462&0.8416&0.8320&0.6607&0.7083&0.6925&0.5353\\       

		& SGH & 0.9336&0.9749&0.7961&0.8816&0.8776&0.7217&0.9178&0.9294&0.7627 \\ \hline     
		
		\multirow{2}{*}{ResNet50~\cite{resnet50}}  & Original & 0.8868&0.9203&0.7240&0.9110&0.9046&0.7533&0.8294&0.8536&0.6591 \\     

		& SGH & 0.9530&0.9749&0.8443&0.9432&0.9273&0.8216&0.9268&0.9358&0.7710\\ \hline     		
		
		\multirow{2}{*}{DeepSRQ~\cite{deepSRQ}}  & Original & 0.9118&0.9428&0.7540&0.7378&0.7358&0.5436&0.7567&0.7961&0.5828\\   
   
		& SGH & 0.9560&0.9729&0.8580&0.7861&0.7628&0.6330&0.9137&0.9256&0.7756\\ \hline     		
		
		\multirow{2}{*}{HyperIQA~\cite{su2020blindly}}  & Original & 0.9085&0.9365&0.7541&0.8931&0.8904&0.7378&0.7901&0.8093&0.6264\\     
  
		& SGH & 0.9470&0.9745&0.8239&0.9356&0.9338&0.7967&0.9246&0.9385&0.7681\\ \hline 
		\multirow{1}{*}{Average Improvement} & - & 0.0526&0.0636&0.0832&0.0473&0.0502&0.0733&0.1660&0.1662&0.1822\\ 
		\bottomrule
	\end{tabular}
\end{table*}

\begin{table*}[!tbhp]\small
	\renewcommand\arraystretch{1.1}
	\centering
	\caption{Cross-dataset performance results of Scale guided hypernetwork (SGH) framework.}
	\label{tab:crossdataset}
	\begin{tabular}{l|c|cc|cc|cc}
		\toprule
		\multirow{2}{*}{Base Model}                                                                  & \multirow{2}{*}{Framework} & \multicolumn{2}{c|}{Waterloo}    & \multicolumn{2}{c|}{QADS} & \multicolumn{2}{c}{CVIU} \\ \cline{3-8} 
		&      & CVIU   &  QADS & CVIU   &   Waterloo     &      Waterloo     & QADS            \\ \hline
		
		\multirow{2}{*}{CNNIQA~\cite{CNNIQA}}  & Original & 0.6478&0.6484&0.6852&0.8552&0.8176&0.7078\\ 
   
		& SGH & 0.8885&0.8253&0.8792&0.8991&0.9314&0.7620\\ \hline     
		
		\multirow{2}{*}{JCSAN~\cite{JCSAN}}  & Original & 0.6326&0.6323&0.7044&0.8613&0.7783&0.6179 \\     
 
		& SGH & 0.9007&0.7927&0.8809&0.9342&0.9401&0.7702\\ \hline  
		
		\multirow{2}{*}{ResNet50~\cite{resnet50}}  & Original & 0.6545&0.5539&0.8121&0.8666&0.9175&0.8474 \\  
  		    
		& SGH & 0.8963&0.7574&0.8884&0.9156&0.9296&0.7976\\ \hline    
		
		\multirow{2}{*}{DeepSRQ~\cite{deepSRQ}}  & Original & 0.7065&0.6686&0.7064&0.8410&0.7390&0.4953 \\      

		& SGH & 0.8932&0.7478&0.8914&0.9395&0.9386&0.7478\\ \hline    
		
		\multirow{2}{*}{HyperIQA~\cite{su2020blindly}}  & Original & 0.7127&0.6863&0.7581&0.8702&0.8557&0.8364 \\      
		& SGH & 0.8758&0.7773&0.8607&0.9189&0.9363&0.7921\\ \hline    

		\multirow{1}{*}{Average Improvement}    

		& - & 0.2201&0.1422&0.1469&0.0626&0.1136&0.0730\\ 
		\bottomrule
	\end{tabular}
\end{table*}

\section{Experiment}
\subsection{Dataset and Evaluation Criterion}
We conduct experiments on three widely-used SR IQA datasets, i.e., Waterloo~\cite{yeganeh2015objective}, CVIU~\cite{ma2017learning}, and QADS~\cite{zhou2019visual} datasets. The details of these datasets are reported in Table~\ref{tbl:dataset details}. We choose three evaluation criteria, i.e., Spearman Rank Correlation Coefficient (SRCC), Pearson Linear Correlation Coefficient (PLCC),
and Kendall’s Rank Correlation Coefficient (KRCC), to quantify the performance of SR IQA metrics. SRCC
and KRCC reflect the prediction monotonicity, while PLCC
measures the prediction accuracy. The higher the value of these evaluation criteria, the better the SR IQA methods. Notice that we calculate SRCC, PLCC, and KRCC through the open-source tool called SciPy~\cite{virtanen2020scipy}.

\subsection{Implementation Details}
We implement the proposed SGH framework as follows. In the content perception stage, the encoder could be seamlessly replaced by feature extractors of existing IQA metrics. In the evaluation rule generation stage, we obtain the scale representation through two FC layers with 128 neurons. In the quality prediction stage, following HyperIQA~\cite{su2020blindly}, we use five FC layers, whose output dimensions are 128, 64, 32, 16, and 1.  The activation function of the first four FC layers is Sigmoid, while that of the last FC layer is the identity function. The hypernetwork generates parameters for five FC layers through five independent modules, where each module uses one FC layer for weights prediction and another for bias prediction. 

Next, we detail the training and test protocols as follows. We randomly split each dataset into the training set and the test set with a ratio of 8:2 while ensuring the image with the same content or the same SR method falls into the same set. The partitioning and evaluation process is repeated
10 times for a fair comparison while considering the
computational complexity, and the average result is reported
as the final performance. At each step of one training epoch, we randomly crop 64 $224\times224$ patches from SR images and feed them into IQA models, whose parameters are optimized through Adam with a learning rate of 1e-4. The training procedure lasts for 50 epochs. At the stage of evaluation,  we first split the input SR image into $224\times224$ patches with a stride of 64, then predict the quality score of all patches, and finally rate the input image the average quality score of all patches.

\subsection{Effectiveness of Scale Guided Hypernetwork Framework}
To verify the effectiveness of the proposed framework, we apply it to five representative blind IQA methods: CNNIQA~\cite{CNNIQA}, ResNet50~\cite{resnet50} based IQA metric, HyperIQA~\cite{su2020blindly}, JCSAN~\cite{JCSAN}, and DeepSRQ~\cite{deepSRQ}. The single-dataset and cross-dataset comparisons between the SGH-enhanced IQA metrics with the original ones are presented in Table~\ref{tab:basemodel} and Table~\ref{tab:crossdataset}, respectively.  From these two tables, several interesting findings can be discovered:

 First, although five IQA models largely differ in network architectures, their single-dataset and cross-dataset performance are significantly boosted by the proposed SGH framework. For example, the average single-dataset improvement of SRCC, PLCC, and KRCC on the CVIU dataset are 0.1660, 0.1662, and 0.1822, respectively. This validates the effectiveness of the proposed SGH framework and also reminds us of the importance of scale information to the task of NR SR IQA. 
 
 Second, the hypernetwork-based IQA metric, i.e., HyperIQA, fails to perform well across different datasets. For instance, although HyperIQA achieves better performance on the Waterloo dataset, it is inferior to ResNet50 on the QADS and CVIU datasets. This indicates that the marriage of hypernetwork and blind SR IQA requires task-specific knowledge. To meet that requirement, we propose a scale guided hypernetwork framework.
 
 Third, the average single-dataset improvement on the CVIU dataset is higher than that on the other two datasets. This is because the statistical effect of scale factors on the visual quality of SR images is more significant on the CVIU dataset, which is derived from Table~\ref{tbl:A-test} where the p-value of the CVIU dataset is lower than that of the other two datasets. 
 
 Fourth, the performance of cross-dataset evaluation on the CVIU dataset is higher than that on the other two datasets. This is because the CVIU dataset contains more images and more scale factors (as shown in Table~\ref{tbl:dataset details}).  

\begin{figure*}[!htbp]
	\centering
	\begin{tabular}{cc}
		\includegraphics[width=0.4\linewidth]{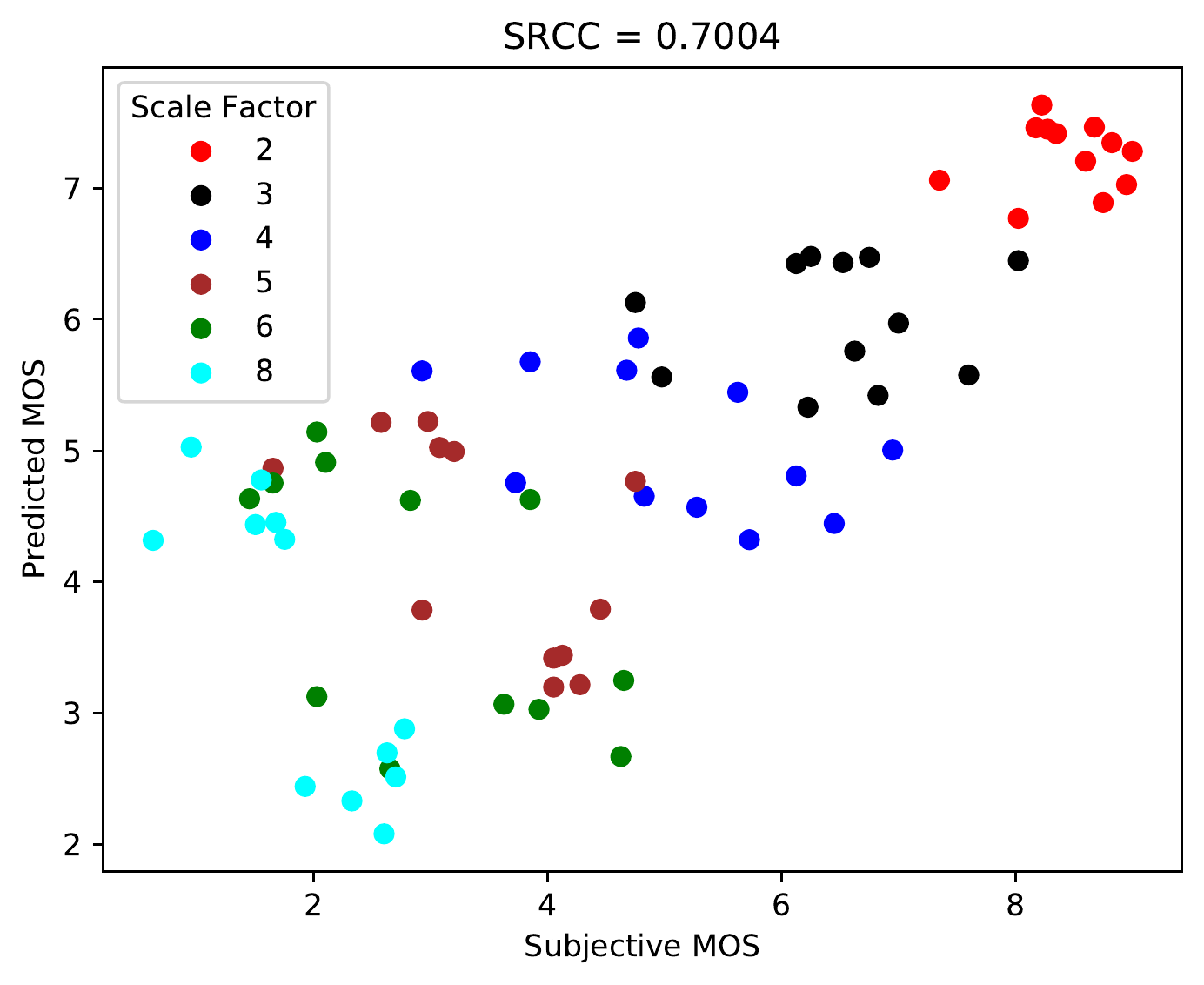} & \includegraphics[width=0.4\linewidth]{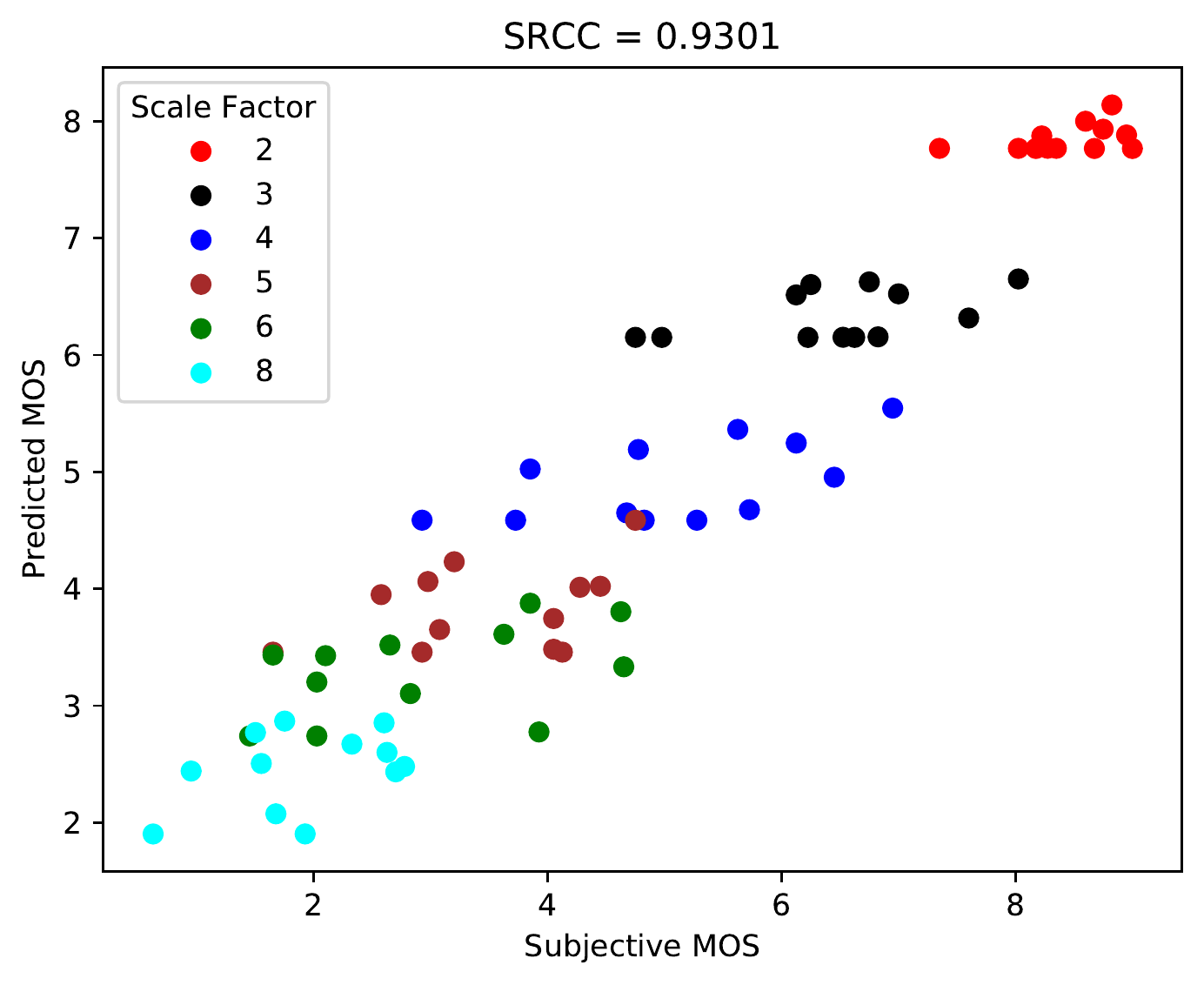} \\ 
		(a) & (b)
	\end{tabular}
	\caption{The scatter plots of predicted MOSs against subjective MOSs on the CVIU dataset. (a) ResNet50; (b) The SGH-enhanced ResNet50. }
	\label{fig:predictedscore}
\end{figure*}

\begin{figure*}[!htbp]
	\centering
	\begin{tabular}{ccc}
		\includegraphics[width=0.25\linewidth]{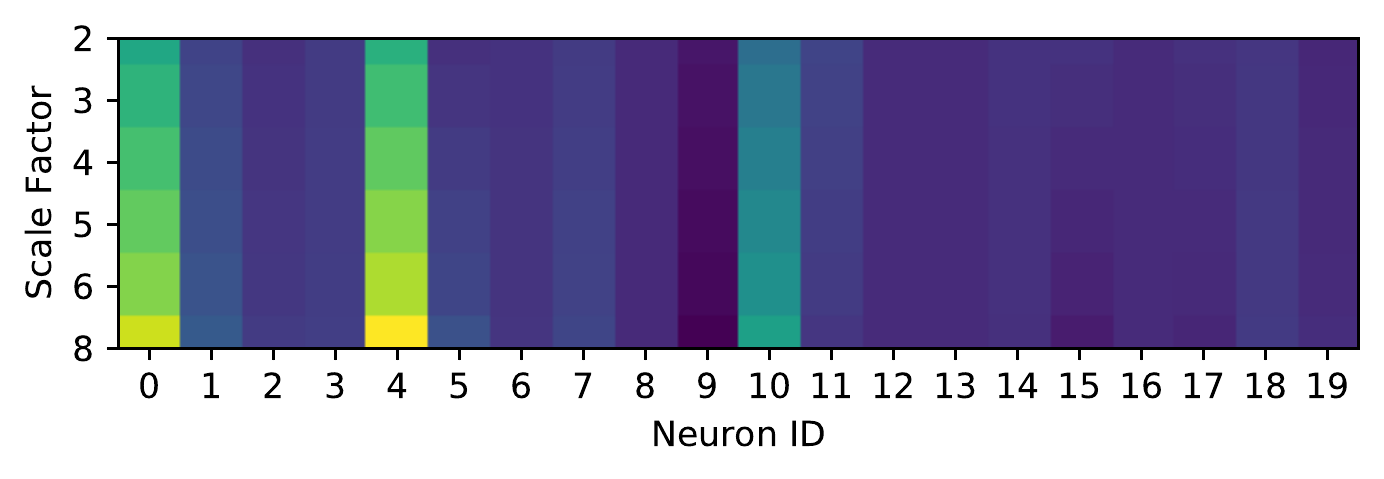} & \includegraphics[width=0.25\linewidth]{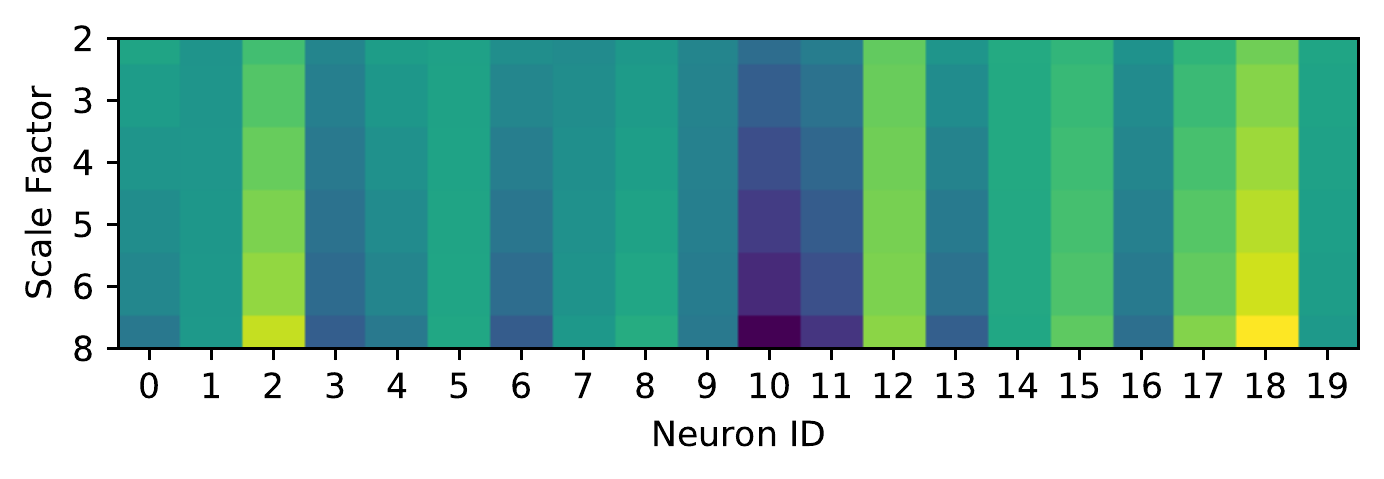} & \includegraphics[width=0.25\linewidth]{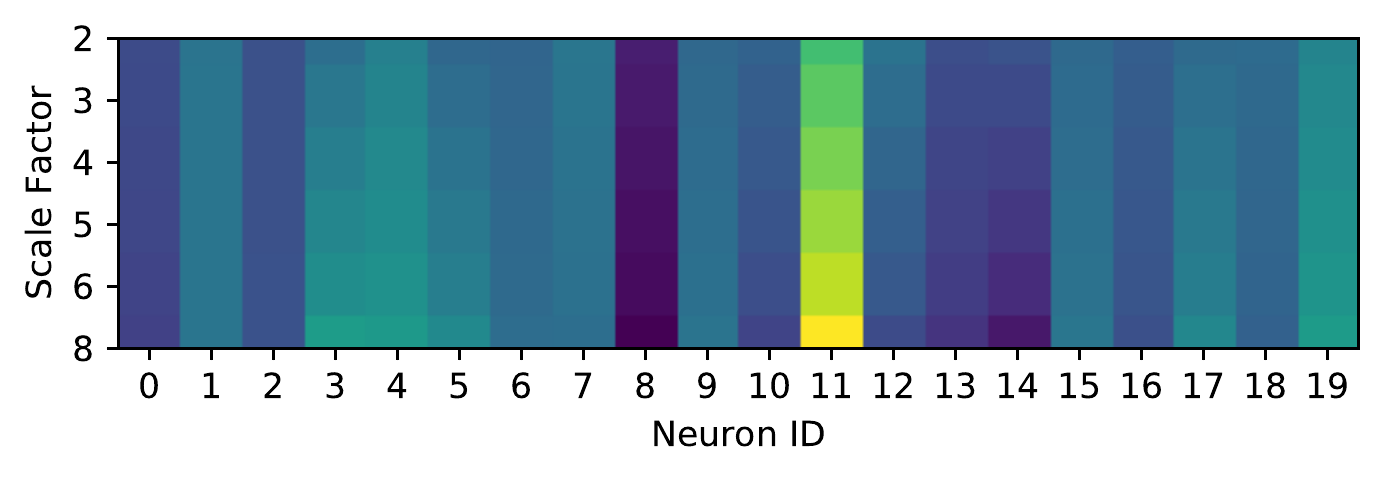} \\ 
		& (a) & \\
		\includegraphics[width=0.25\linewidth]{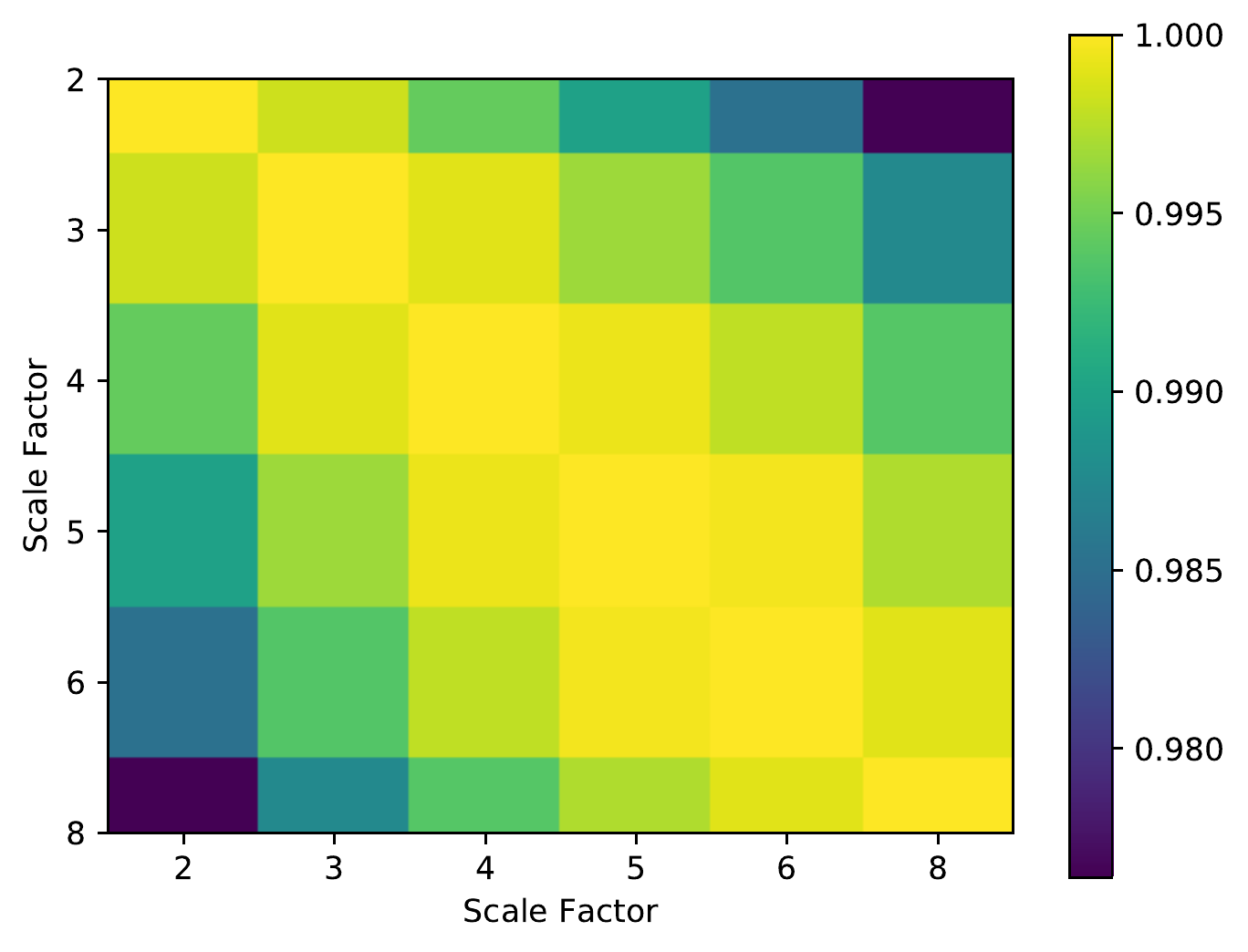} & \includegraphics[width=0.25\linewidth]{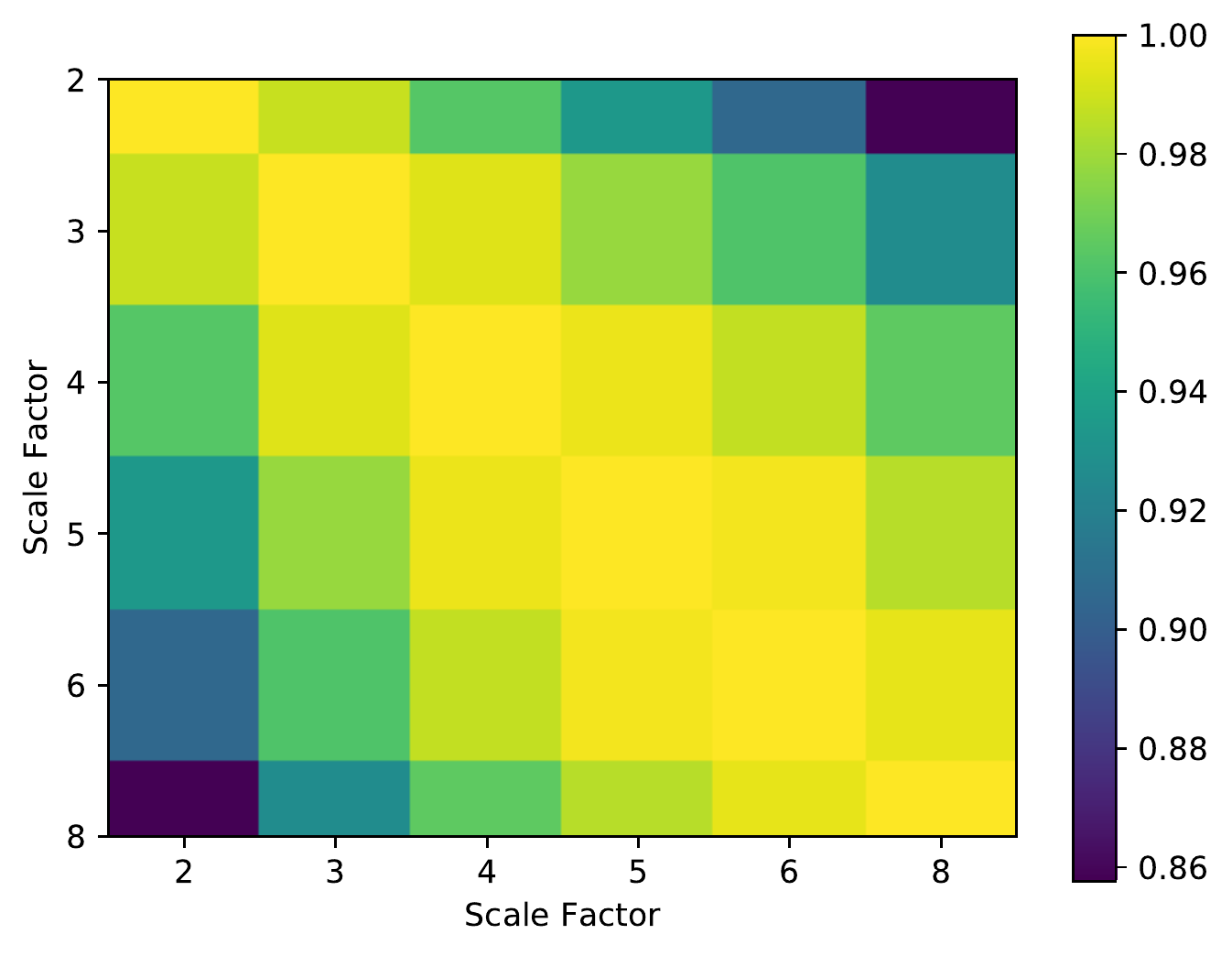} & \includegraphics[width=0.25\linewidth]{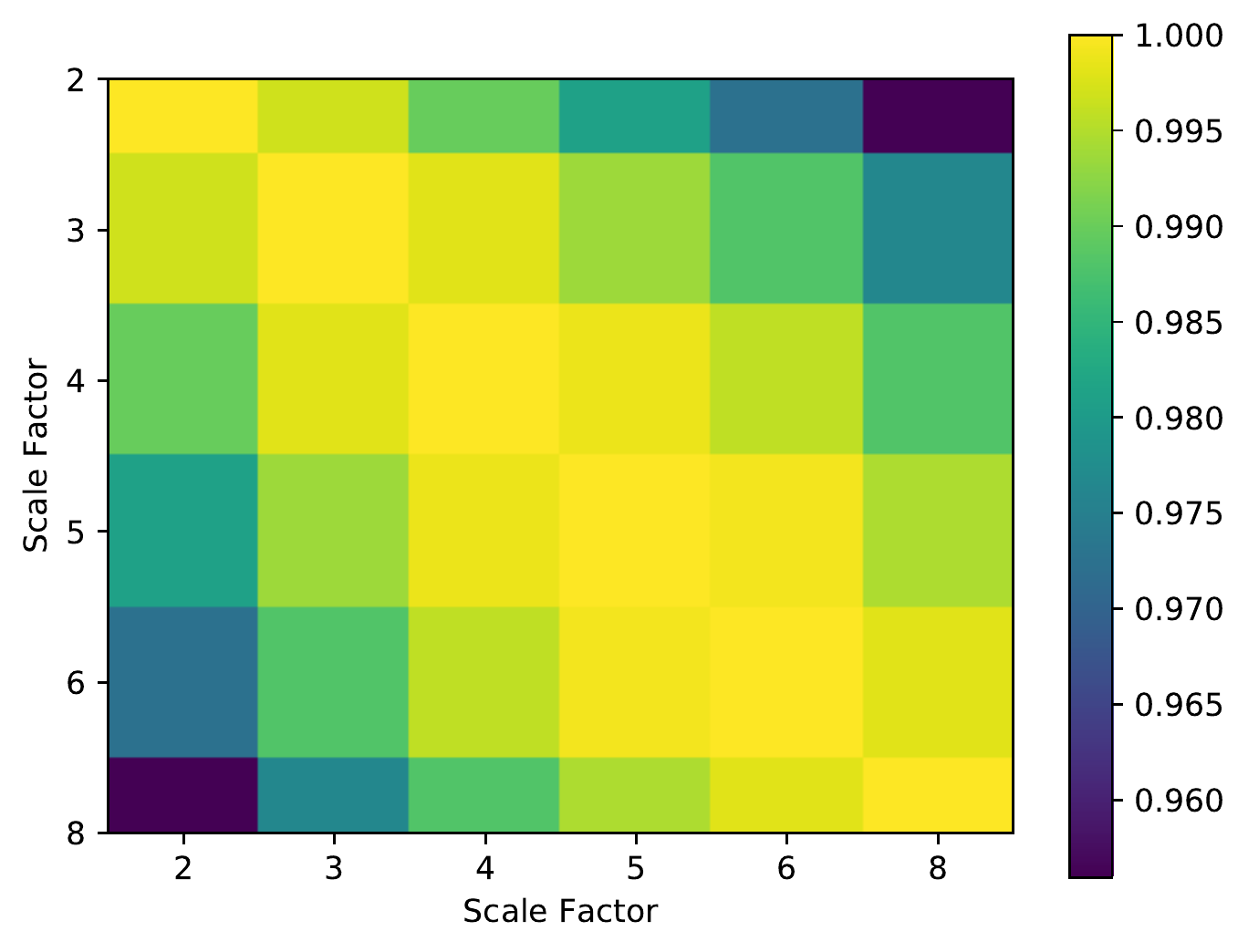} \\
		& (b) & \\ 
	\end{tabular}
	\caption{Generated weights of the first three FC layers and their correlation maps. (a) The first 20 neurons of weights. (b) The cosine similarity between generated weights. }
	\label{fig:vis1}
\end{figure*}
\subsection{Deep Dive to the Scale Guided Hypernetwok}
To better understand the performance gains obtained by SGH, we plot predicted MOS against subjective MOS in Fig.~\ref{fig:predictedscore}. As shown, a large portion of SGH’s SRCC gains come from its ability to limit errors in assigning higher quality scores to SR images with larger scale factors. This is thanks to the scale priori, that is, SR images with lower scale factors are generally more visually pleasing than those with high scale factors.

In order to understand how the scale priori helps us assess image quality, we visualize the generated weights of FC layers used in quality predictions and the similarity between weights of different scale factors in Fig.~\ref{fig:vis1}. As shown, several interesting findings can be discovered:

First, for different scale factors, the generated weights vary. This indicates that our framework adopts a unique evaluation rule for each scale factor. Whereas for traditional IQA models, the evaluation rule is fixed for all scale factors. 

Second, weights generated for different scale factors are highly correlated. This indicates that our framework slightly adjusts the evaluation rule for SR images according to the scale factor. This makes sense because the appearance of the image does not decline dramatically as the scale factor slightly increases.

\subsection{Comparison with the State-of-the-art}
To demonstrate the advantage of the SGH-enhanced IQA model, we make a comparison with state-of-the-art IQA methods including 11 FR IQA models~\cite{ssim,msssim,iwssim,fsim,gmsd,msgmsd,vsi,mdsi,haarpsi,sfsn} and 7 NR IQA approaches~\cite{brisque,dbcnn,CNNIQA,JCSAN,deepSRQ,resnet50,su2020blindly}. The performance of works~\cite{ssim,msssim,iwssim,fsim,gmsd,msgmsd,vsi,mdsi,haarpsi,brisque} is calculated by the image quality assessment toolbox named PyIQA~\cite{pyiqa}, while that of the remaining methods are based on their source codes. The mean results over 10 trials are reported in Table~\ref{tab:result}. From this table, we observe that the SGH-enhanced ResNet50 achieves the best performance across three SR IQA datasets. This verifies the advantage of the proposed SGH framework. 

\begin{figure*}[htbp]
	\centering
	\begin{tabular}{ccc}
		\includegraphics[height=3.4cm,width=0.3\linewidth]{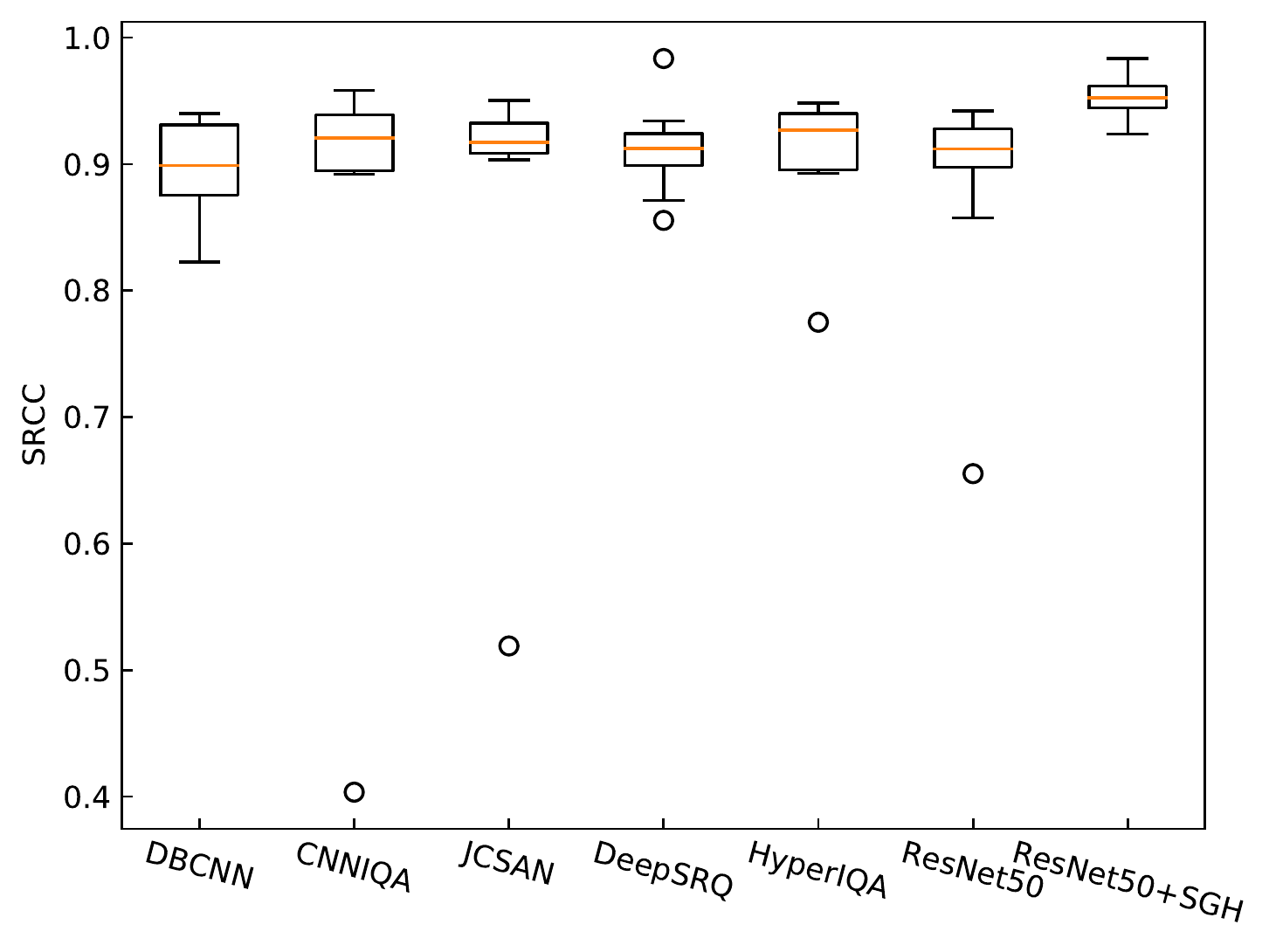} & \includegraphics[height=3.4cm,width=0.3\linewidth]{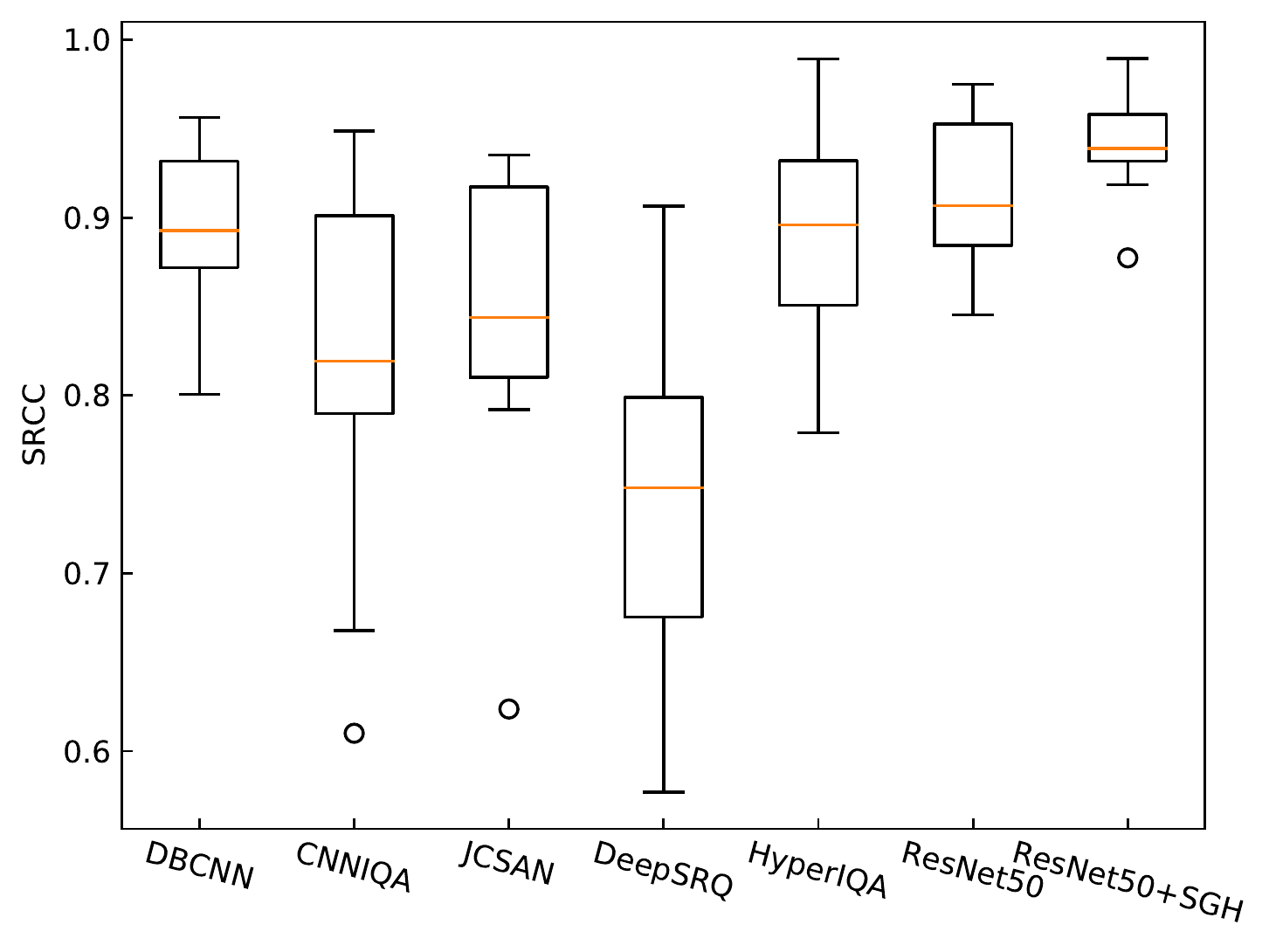} & \includegraphics[height=3.4cm,width=0.3\linewidth]{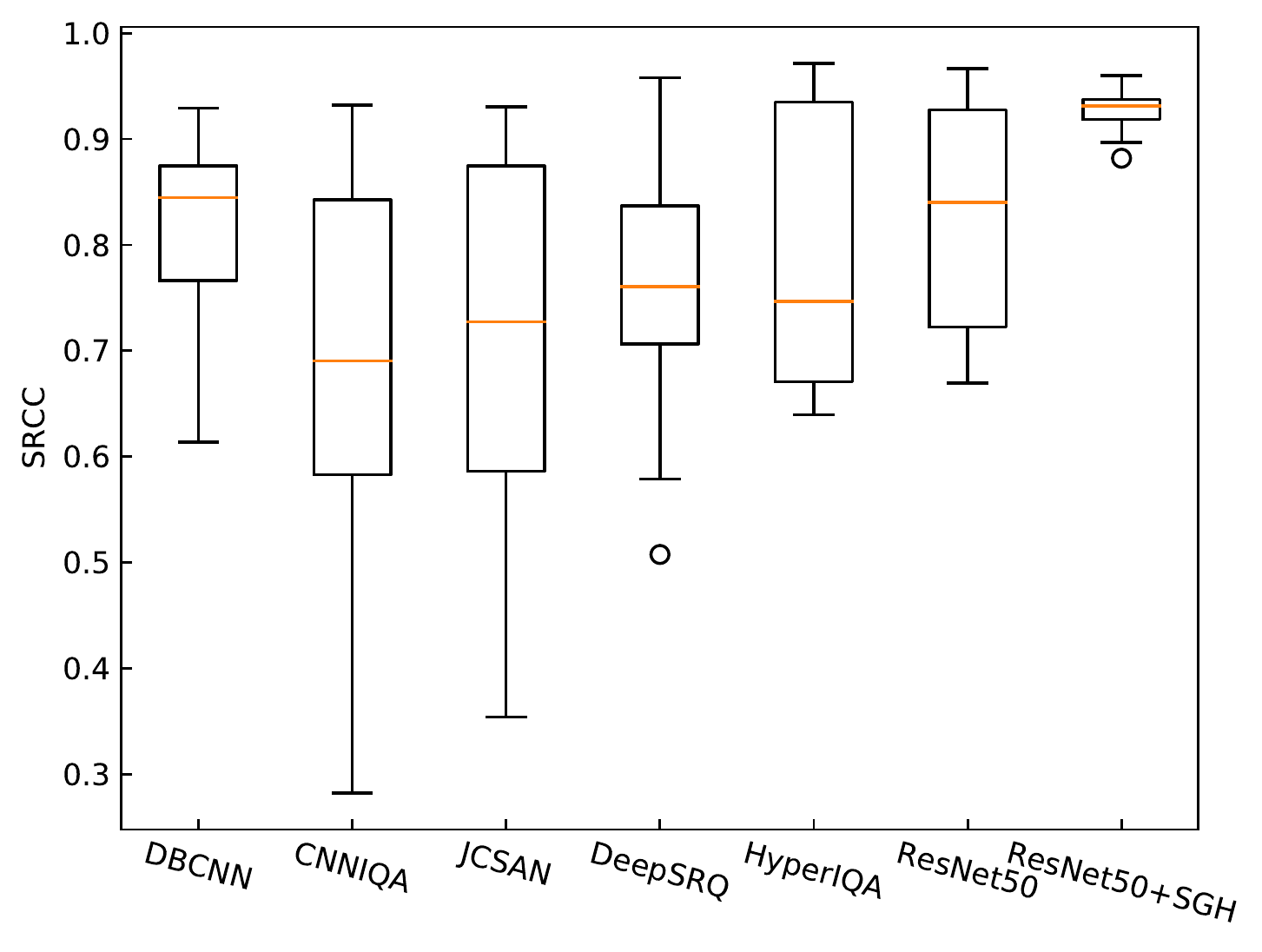} \\ 
		(a) & (b) & (c)\\
	\end{tabular}
	\caption{Box plot of SRCC distributions of IQA models over 10 trials. (a) On the Waterloo dataset; (b) On the QADS dataset; (c) On the CVIU dataset.}
	\label{fig:boxplot}
\end{figure*}

To further analyze the performance stability of the SGH enhanced IQA model, we plot SRCC distributions of IQA models over 10 trials. As shown in Fig.~\ref{fig:boxplot}, the performance results of the SGH-enhanced IQA model are more concentrated than those of other 6 NR IQA models, indicating that the SGH-enhanced IQA model has better stability.

\begin{table*}[!tbhp]\small
	\renewcommand\arraystretch{1.1}
	\centering
	\caption{Comparison with state-of-the-art FR and NR IQA models. The best performance results are bold.}
	\label{tab:result}
	\begin{tabular}{l|l|ccc|ccc|ccc}
		\toprule
		\multirow{2}{*}{Type}                                                                  & \multirow{2}{*}{Metric} & \multicolumn{3}{c|}{Waterloo}    & \multicolumn{3}{c|}{QADS} & \multicolumn{3}{c}{CVIU} \\ \cline{3-11} 
		&      & SRCC   &  PLCC & KRCC    & SRCC        &  PLCC & KRCC          & SRCC       &  PLCC & KRCC     \\ \hline
		\multirow{11}{*}{FR IQA}         &     PSNR &0.8163& 0.8107& 0.6442& 0.4229& 0.4052& 0.3246& 0.6157& 0.6220& 0.4464\\
		&SSIM~\cite{ssim}& 0.9085& 0.9143& 0.7567& 0.5864& 0.5836& 0.4452& 0.6339& 0.6372& 0.4551\\
		&MS-SSIM~\cite{msssim}& 0.9023& 0.8952& 0.7510& 0.7402& 0.6988& 0.5657& 0.7800& 0.7406& 0.5897\\
		&IW-SSIM~\cite{iwssim}& 0.9161& 0.9318& 0.7641& 0.8144& 0.7814& 0.6385& 0.8443& 0.8144& 0.6605\\
		&FSIM~\cite{fsim}& 0.9063& 0.9216& 0.7461& 0.6882& 0.6779& 0.5200& 0.7559& 0.7550& 0.5647\\
		&GMSD~\cite{gmsd}& 0.8797& 0.9080& 0.7047& 0.7791& 0.7656& 0.6044& 0.8480& 0.8535& 0.6677\\
		&MS-GMSD~\cite{msgmsd}& 0.8808& 0.9131& 0.7065& 0.7457& 0.7322& 0.5684& 0.8384& 0.8430& 0.6554\\
		&VSI~\cite{vsi}& 0.8752& 0.8911& 0.7065& 0.5964& 0.5770& 0.4415& 0.7341& 0.7208& 0.5519\\
		&MDSI~\cite{mdsi}& 0.9090& 0.9157& 0.7518& 0.6345& 0.6294& 0.4686& 0.7848& 0.7904& 0.5952\\
		&HaarPSI~\cite{haarpsi}& 0.8998& 0.9387& 0.7467& 0.8028& 0.8131& 0.6287& 0.8386& 0.8506& 0.6558\\
		
		& SFSN~\cite{sfsn} &0.8937& 0.9193& 0.7185& 0.7980& 0.7725& 0.6258& 0.8398& 0.8109& 0.6543 \\\hline
		\multirow{8}{*}{NR IQA} 
		& BRISQUE~\cite{brisque} & 0.8297& 0.8390& 0.6283& 0.4925& 0.5060& 0.3543& 0.6676& 0.7033& 0.4984\\
		& DBCNN~\cite{dbcnn}& 0.8969&0.9146&0.7174&0.8921&0.8959&0.7331&0.8154&0.8510&0.6359 \\
		& CNNIQA~\cite{CNNIQA}   & 0.8717&0.8829&0.7318&0.8169&0.7756&0.6411&0.6789&0.6750&0.5154    \\
		& JCSAN~\cite{JCSAN} & 0.8839&0.8696&0.7462&0.8416&0.8320&0.6607&0.7083&0.6925&0.5353 \\
		& DeepSRQ~\cite{deepSRQ} & 0.9118&0.9428&0.7540&0.7378&0.7358&0.5436&0.7567&0.7961&0.5828\\
		& ResNet50~\cite{resnet50} & 0.8868&0.9203&0.7240&0.9110&0.9046&0.7533&0.8294&0.8536&0.6591 \\		
		& HyperIQA~\cite{su2020blindly} &  0.9085&0.9365&0.7541&0.8931&0.8904&0.7378&0.7901&0.8093&0.6264\\
		& ResNet50 + SGH & \textbf{0.9530}&\textbf{0.9749}&\textbf{0.8443}&\textbf{0.9432}&\textbf{0.9273}&\textbf{0.8216}&\textbf{0.9268}&\textbf{0.9358}&\textbf{0.7710 }\\  \bottomrule
	\end{tabular}
\end{table*}

\begin{table*}[!tbhp]\small
	\renewcommand\arraystretch{1.1}
	\centering
	\caption{Single-dataset comparison with the fusion-based framework where we predict the quality score based on the fused representation of scale factor and image features.}
	\label{tab:ablationstudy}
	\begin{tabular}{l|c|ccc|ccc|ccc}
		\toprule
		\multirow{2}{*}{Base Model}                                                                  & \multirow{2}{*}{Framework} & \multicolumn{3}{c|}{Waterloo}    & \multicolumn{3}{c|}{QADS} & \multicolumn{3}{c}{CVIU} \\ \cline{3-11} 
		&      & SRCC   &  PLCC & KRCC    & SRCC        &  PLCC & KRCC          & SRCC       &  PLCC & KRCC     \\ \hline
		
		\multirow{3}{*}{CNNIQA~\cite{CNNIQA}}  & Original & 0.8717&0.8829&0.7318&0.8169&0.7756&0.6411&0.6789&0.6750&0.5154\\      
		& Fusion-based & 0.9321&0.9698&0.7921&0.8927&0.8818&0.7268&0.8994&0.9115&0.7366\\    
 
		& SGH  & 0.9260&0.9719&0.7815&0.8711&0.8674&0.6990&0.8936&0.9120&0.7274      \\ \hline     
		\multirow{3}{*}{JCSAN~\cite{JCSAN}}  & Original & 0.8839&0.8696&0.7462&0.8416&0.8320&0.6607&0.7083&0.6925&0.5353\\      
		& Fusion-based  & 0.9311&0.9593&0.7986&0.8927&0.8875&0.7245&0.9077&0.9198&0.7454\\    
  
		& SGH  &0.9275&0.9735&0.7842&0.8983&0.8946&0.7384&0.9092&0.9264&0.7468		\\ \hline
		\multirow{3}{*}{ResNet50~\cite{resnet50}}  & Original & 0.8868&0.9203&0.7240&0.9110&0.9046&0.7533&0.8294&0.8536&0.6591 \\      
		& Fusion-based  & 0.8992&0.9326&0.7502&0.9308&0.9238&0.7894&0.8522&0.8758&0.6816\\    

				& SGH &  0.9067&0.9671&0.7383&0.9445&0.9416&0.8164&0.9013&0.9216&0.7386\\ \hline   
		\multirow{3}{*}{DeepSRQ~\cite{deepSRQ}}  & Original & 0.9118&0.9428&0.7540&0.7378&0.7358&0.5436&0.7567&0.7961&0.5828\\      
		& Fusion-based  & 0.9129&0.9534&0.7591&0.8431&0.8365&0.6647&0.9140&0.9113&0.7519\\    
 		  		
		& SGH &  0.9180&0.9734&0.7593&0.8673&0.8600&0.6901&0.9158&0.9328&0.7597\\ \hline       
		
		\multirow{3}{*}{HyperIQA~\cite{su2020blindly}}  & Original & 0.9085&0.9365&0.7541&0.8931&0.8904&0.7378&0.7901&0.8093&0.6264\\      
		& Fusion-based  & 0.8927&0.9347&0.7423&0.9027&0.8878&0.7471&0.7941&0.8129&0.6308 \\    

		& SGH &  0.9247&0.9674&0.7763&0.9351&0.9305&0.8013&0.9121&0.9262&0.7533\\ \hline 
		\multirow{2}{*}{Average Improvement}    
		& Fusion-based  & 0.0211&0.0395&0.0264&0.0523&0.0558&0.0632&0.1208&0.1210&0.1255\\    
 & SGH & 0.0526&0.0636&0.0832&0.0473&0.0502&0.0733&0.1660&0.1662&0.1822\\ 

		\bottomrule
	\end{tabular}
\end{table*}

\begin{table*}[!tbhp]\small
	\renewcommand\arraystretch{1.1}
	\centering
	\caption{Cross-dataset comparison with the fusion-based framework.}
	\label{tab:crossdataset2}
	\begin{tabular}{l|l|cc|cc|cc}
		\toprule
		\multirow{2}{*}{Base Model}                                                                  & \multirow{2}{*}{Framework} & \multicolumn{2}{c|}{Waterloo}    & \multicolumn{2}{c|}{QADS} & \multicolumn{2}{c}{CVIU} \\ \cline{3-8} 
		&     &  CVIU   &  QADS & CVIU   &   Waterloo     &      Waterloo     & QADS            \\ \hline
		
		\multirow{3}{*}{CNNIQA~\cite{CNNIQA}}  & Original & 0.6478&0.6484&0.6852&0.8552&0.8176&0.7078\\      
		& Fusion-based & 0.9000&0.8295&0.8605&0.9253&0.9379&0.7621\\    

		& SGH & 0.8885&0.8253&0.8792&0.8991&0.9314&0.7620\\ \hline     
				
		\multirow{3}{*}{JCSAN~\cite{JCSAN}}  & Original & 0.6326&0.6323&0.7044&0.8613&0.7783&0.6179 \\      
		& Fusion-based & 0.9014&0.8200&0.8698&0.9277&0.9355&0.7760\\    

		& SGH & 0.9007&0.7927&0.8809&0.9342&0.9401&0.7702\\ \hline  
				
		\multirow{3}{*}{ResNet50~\cite{resnet50}}  & Original & 0.6545&0.5539&0.8121&0.8666&0.9175&0.8474 \\      
		& Fusion-based & 0.6663&0.6293&0.8408&0.8923&0.9016&0.8392\\    

				& SGH & 0.8963&0.7574&0.8884&0.9156&0.9296&0.7976\\ \hline    
				
		\multirow{3}{*}{DeepSRQ~\cite{deepSRQ}}  & Original & 0.7065&0.6686&0.7064&0.8410&0.7390&0.4953 \\      
		& Fusion-based & 0.8800&0.7225&0.8700&0.9059&0.9201&0.7100\\    

			& SGH & 0.8932&0.7478&0.8914&0.9395&0.9386&0.7478\\ \hline    
				
		\multirow{3}{*}{HyperIQA~\cite{su2020blindly}}  & Original & 0.7127&0.6863&0.7581&0.8702&0.8557&0.8364 \\      
		& Fusion-based & 0.6876&0.6540&0.7973&0.8682&0.8737&0.8340\\    

		& SGH & 0.8758&0.7773&0.8607&0.9189&0.9363&0.7921\\ \hline   
		\multirow{2}{*}{Average Improvement}    
		& Fusion-based & 0.1362&0.0932&0.1144&0.0450&0.0921&0.0833\\    
		& SGH &  0.2201&0.1422&0.1469&0.0626&0.1136&0.0730\\ 
		\bottomrule
	\end{tabular}
\end{table*}

\subsection{Ablation study}
To verify the superiority of the proposed scale guided hypernetwork (SGH) framework in exploiting scale information, we compare it with the fusion-based framework where we directly concatenate scale information and image features for quality score prediction. The results of the single-dataset comparison are presented in Table~\ref{tab:ablationstudy}. According to Table~\ref{tab:ablationstudy}, we have the following findings:

 First, the fusion-based framework significantly improves the performance of five IQA models, which indicates that direct concatenation is a simple yet efficient approach to using scale information to guide the task of SR IQA. 
 
 Second, the average improvement brought by the proposed SGH is higher than that brought by the fusion-based framework. This also occurs in Table~\ref{tab:crossdataset2}, which reports the results of the cross-dataset comparison. Therefore, we can conclude that the proposed SGH framework is a more effective way to exploit scale information to guide blind SR IQA than the fusion-based one.

\section{Conclusion}
In this paper, we first reveal that the scale factor can be used to guide the task of blind SR IQA through qualitative and quantitative experiments, and then propose a scale guided hypernetwork framework. In the proposed framework, we evaluate SR image quality in a scale-adaptive manner, where the evaluation rule is generated based on the scale factor of the SR image. Experimental results demonstrate that the proposed framework not only improves the performance of existing IQA metrics but also enhances their generalization abilities.

\section*{Acknowledgments}
Thanks to Dr. Jiachen Gu for providing computing resources. 

\bibliographystyle{IEEEtran}
\bibliography{sgh}

\end{document}